# SUPPORT VECTOR MACHINE FOR DETERMINING EULER ANGLES IN AN INERTIAL NAVIGATION SYSTEM

A.N. Grekov[1,2], A.A. Kabanov[2], S.Yu. Alekseev[1]

[1]Institute of Natural and Technical Systems, RF, Sevastopol, Lenin St., 28
[2]Sevastopol State University, RF, Sevastopol, Universitetskaya St., 33
*E-mail: oceanmhi@ya.ru*

The paper discusses the improvement of the accuracy of an inertial navigation system created on the basis of MEMS sensors using machine learning (ML) methods. As input data for the classifier, we used information obtained from a developed laboratory setup with MEMS sensors on a sealed platform with the ability to adjust its tilt angles. To assess the effectiveness of the models, test curves were constructed with different values of the parameters of these models for each core in the case of a linear, polynomial radial basis function. The inverse regularization parameter was used as a parameter. The proposed algorithm based on MO has demonstrated its ability to correctly classify in the presence of noise typical for MEMS sensors, where good classification results were obtained when choosing the optimal values of hyperparameters.
**Keywords**: navigation, accelerometer, acceleration, gyroscope, magnetometer, error, autonomous underwater vehicles, machine learning, algorithm

**Introduction.** Aquatic monitoring is carried out today by experienced professionals, which is time-consuming and increases operating costs and risks to the health of employees. The use of modern technologies to reform the traditional industry is of great importance to improve the efficiency and quality of such monitoring. With the development of technology, autonomous underwater vehicles (AUVs) gradually began to play a key role in the study of the marine environment. Currently, AUVs are mainly used in geodetic applications in large marine areas, as well as a carrier platform for monitoring water quality parameters in the marine environment using sensors [1, 2].

Navigation is one of the key technologies for AUVs because localization, track tracking and platform control are based on accurate navigational parameters. Some navigation methods commonly used on land and in the air are not suitable for underwater applications due to the attenuating effect of water on electromagnetic signals, so underwater navigation is a major problem with autonomous submersibles. The inertial navigation system (INS), using, for example, microelectromechanical sensors (MEMS) [3], acts as the main AUV navigation system.

As a rule, an INS consists of accelerometers that measure linear acceleration, gyroscopes that measure angular velocity, and magnetometers that measure the magnetic field. For inertial navigation, the instantaneous speed and position of the underwater platform are obtained by integrating the measured values of accelerometers, gyroscopes, and magnetometers [4]. INS errors increase with the elapsed time due to the drift of accelerometers and gyroscopes. Theoretically, speed and heading errors accumulate linearly over time, while position error accumulates exponentially over time [5]. Thus, an inertial navigation system can provide relatively accurate navigation information for a short time, but it is physically impossible for an INS to maintain a high level of accuracy throughout the entire movement. Auxiliary sensors or other navigation systems such as Doppler log (DVL), pressure sensor, global positioning system (GPS), acoustic positioning system (APS) or geophysical navigation system are combined with the INS to form an integrated navigation system to improve positioning accuracy [6]. Autonomous underwater vehicles with multi-parametric water quality sensors allow you to control the parameters of the marine environment in accordance with the planned routes. To achieve the above functions, the AUV must have an accurate navigation and positioning system. In our work, to improve the accuracy of ANNs based on MEMS sensors, we used the methods of Machine Learning (ML) [7].

Recently, MO methods have been actively used to solve navigation problems, for example, when navigating indoors [8–10]. The use of an ensemble of support vector methods was proposed in [11] for the initial adjustment of a strapdown inertial navigation system, where it showed better performance than the traditional approach with Kalman filtering. A deep learning model for determining the instantaneous speed in inertial navigation systems was proposed in [12]. In [13], to eliminate sources of errors in the signals of inertial sensors, the authors applied the convolutional neural network (CNN) method. In another recent paper [14], the authors proposed a navigation method using data from low-cost INS sensors (accelerometers and gyroscopes) on a moving vehicle using machine learning methods instead of the traditional method using Kalman filters.

**Materials and methods.** To solve various problems, many different MO algorithms have been developed [15]. According to D.H. Wolpert "On the absence of free lunches" ("no free lunch" (NFL)) [16], it can be said that no classifier performs better than others in all possible scenarios. In practice, it is always recommended to compare the performance of at least several different learning algorithms in order to choose the best model for the given task. One common metric is classification accuracy, which is defined as the proportion of correctly classified samples. The performance of a classifier—computing power and predictive power—is highly dependent on the underlying data available for training.

As input data for our classifier, we use information obtained from MEMS sensors of the developed navigation platform installed in a sealed container placed in an aquarium with water [6]. The layout of the navigation platform consists of a gyroscope, an accelerometer, a magnetometer and differential pressure sensors, the information from which is fed to the PC. On fig. 1 shows a laboratory setup for research, where to obtain classes, platform inclination angles (from 0° to 30°) in each mutually perpendicular direction are used.

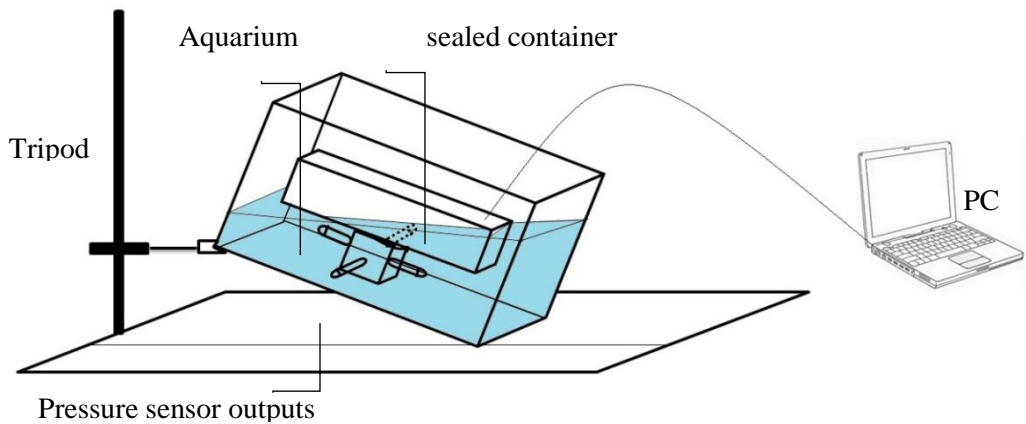

**Fig. 1.** Schematic of a laboratory setup for conducting research with a sealed a navigation platform placed in the water and the ability to adjust the tilt angles

To evaluate how well the trained model performs on data that has not been seen before, we will split the dataset into training and test sets. Using the train_test_split function from the model_selection module of the scikit-learn library [17], we randomly split the X and Y arrays into 30% test data (2893 samples) and 70% training data (6749 samples). The train_test_split function distributes the training sets before splitting into uniform classes between the test and training sets.

To achieve optimal efficiency, many MO algorithms require features to be scaled to the same scale. In our work, we standardized the features [18] using the StandardScaler class from the preprocessing module of the scikit-learn library [17]. The standardization procedure can be expressed using the equation:

$$x_{std}^{(i)} = \frac{x^{(i)} - \mu_x}{\sigma_x} \quad (1)$$

Here $x^i$ is an individual sample, $\mu_x$ is the sample mean for a certain feature column, and $\sigma_x$ is the corresponding standard deviation.

**Support vector machine (SVM).** One of the most powerful and flexible supervised learning algorithms, which can be considered an extension of the perceptron [19]. By applying the perceptron algorithm, we minimize misclassification errors, and in the SVM method, the goal of optimization is to maximize the gap. The gap is defined as the distance between the separating hyperplane (decision boundary) and the training samples closest to this hyperplane, which are called support vectors [20–23]. The original SVM algorithm was proposed by Vapnik and Chervonenkis in a paper on statistical learning in 1963 to find the optimal hyperplane dividing data into two clusters. Nearly thirty years later, Boser, Guyon, and Vapnik developed non-linear classifiers using a nuclear trick to construct a hyperplane with a maximum gap [24]. In [25], Cortes and Vapnik introduced the concept of a dummy variable, which led to the appearance of a modern version of the classifier, which has become standard (with a soft gap). The SVM classifier can be represented as [26]:

$$f(x) = \beta_0 + \sum_{i \in S} \alpha_i K(x_i, x_{i'}), \quad (2)$$

where $B_0$ is the offset; S is the set of all support vector observations; $\alpha_i$ - model parameters that need to be memorized; $x_i$, $x_{i'}$, are pairs of two support vector observations, K is a kernel function that compares $x_i$ and $x_{i'}$ and is a proximity function.

To increase the efficiency of our models, we will construct verification curves for various values of the parameters of these models for each SVM core. One such parameter is the inverse regularization parameter C. Large values of C correspond to large error penalties, while choosing smaller values for C means less stringency for misclassification errors.

In the case of a linear kernel:

$$K(x_i, x_{i'}) = \sum_{j=1}^{p} x_{ij} x_{i'j}, \quad (3)$$

where $p$ is the number of features, the test curve for the C parameter has the form shown in fig. 2. As can be seen from the graph, the model underfits on the data with an increase in the strength of regularization (C < 10). For values of C > 10, the model tends to slightly overfit, as validation accuracy begins to decrease as training accuracy increases. For the linear kernel, the optimal value of C is 10, which allows us to obtain the correctness of the classifier prediction at the level of 0.907.

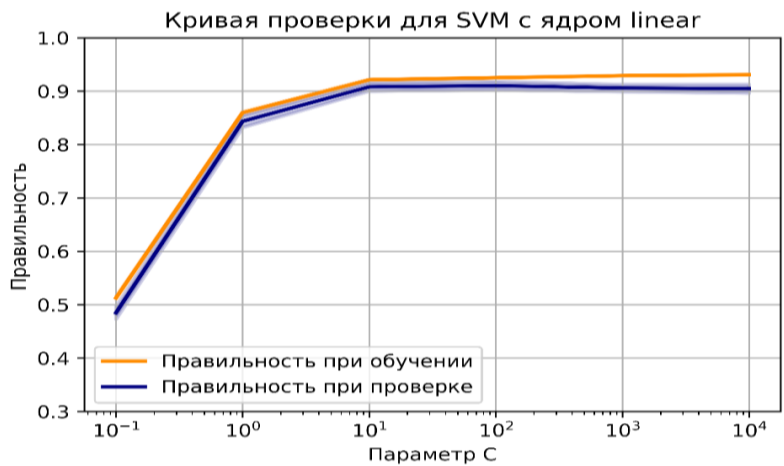

**Fig. 2.** Test curves in the case of a linear kernel

In the case of a polynomial kernel

$$K(x_i, x_{i'}) = \left(1 + \sum_{j=1}^{p} x_{ij} x_{i'j}\right)^2, \quad (4)$$

where n is the degree of the polynomial kernel function, the verification curve for parameter C has the form shown in Fig. 3 with the degree of the polynomial function equal to 3 and the parameter $\gamma = 0.1$. The parameter $\gamma$ can be considered as the cutoff parameter for the Gaussian sphere. If we increase the value of $\gamma$, then we increase the influence or coverage of the training samples, which leads to a denser and bumpier decision boundary. As can be seen from the graph, the model underfits on the data with an increase in the regularization strength (C < 1000). For values of C > 1000, the model tends to slightly overfit the data. For a polynomial kernel, the optimal value for C is 1000.

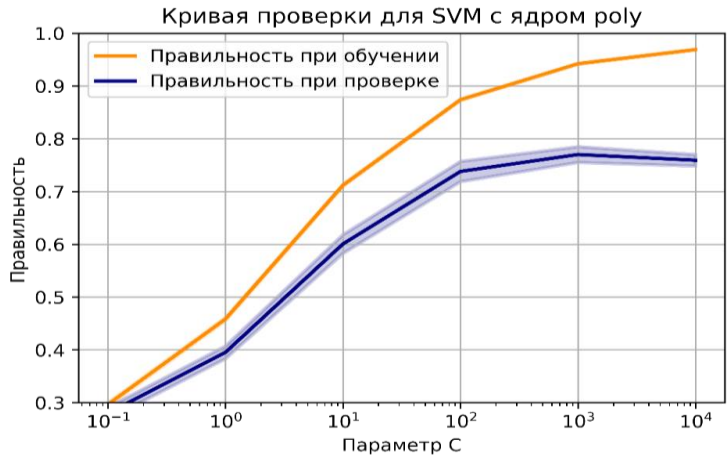

**Fig. 3.** Test curves for a polynomial kernel function with the degree of the polynomial function equal to three

Now let's build a verification curve for the parameter $\gamma$, with the degree of the polynomial function 3 and the optimal parameter C = 1000. From the graph presented in fig. 4, it can be seen that for gamma values less than 0.1, the model is undertrained. Gamma values around 0.1 result in high values for both scores, i.e. the classifier works well. If gamma exceeds 0.1, the classifier is retrained. With optimal values of hyperparameters, the use of a polynomial kernel makes it possible to obtain the correct prediction of the classifier at the level of 0.764.

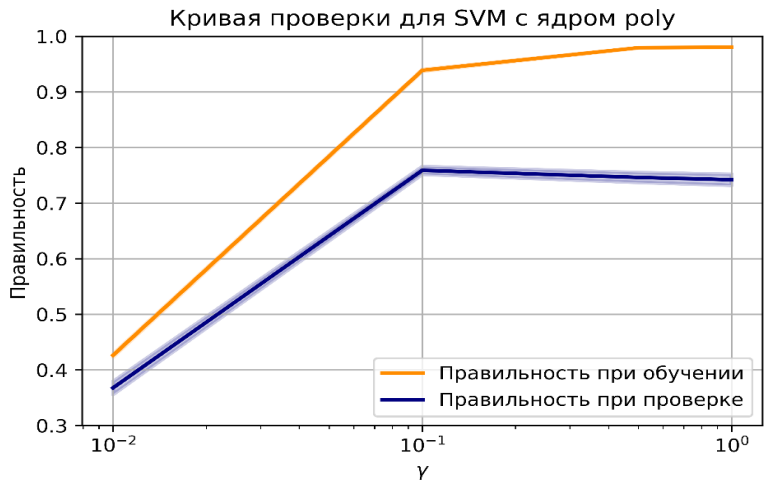

**Fig. 4.** Dependence of the curves of checking the polynomial kernel function on gamma

In the case of the kernel of the radial basis function (RBF):

$$K(x_i, x_{i'}) = \exp\left(-\gamma \sum_{j=1}^{p}(x_{ij}x_{i'j})^2\right), \quad (5)$$

the verification curve for parameter C has the form shown in fig. 5, with the parameter γ = 0.1. As can be seen from the graph, the model underfits on the data with an increase in the strength of regularization (C < 100). For C > 100, the model tends to slightly overfit. For the rbf kernel, the optimal value for C is 100.

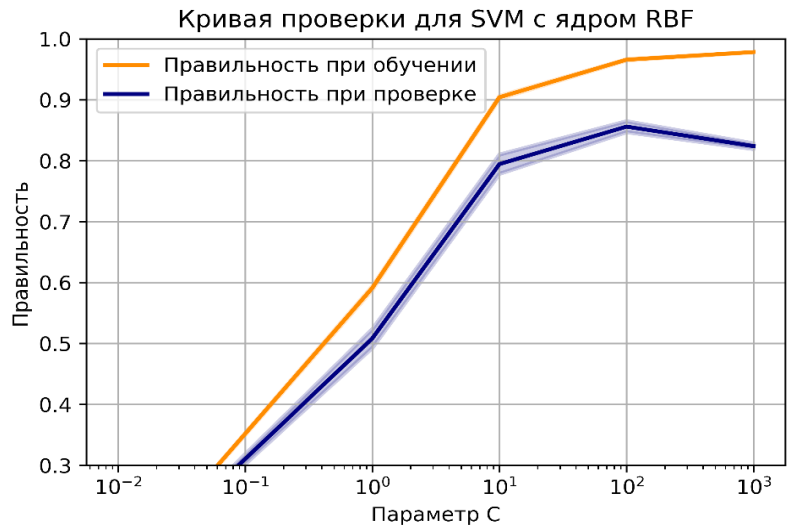

**Fig. 5.** Test curves with radial basis function kernel

Now let's build a verification curve for the parameter γ with the optimal parameter C = 100. From the graph presented in fig. 6, it can be seen that for gamma values less than 0.01, the model is underfit. Gamma values around 0.1 result in high values for both scores, i.e. the classifier works well. If gamma exceeds 0.01, the classifier is retrained. For the obtained optimal values of hyperparameters for the SBF core, the correctness of the classifier is 0.902.

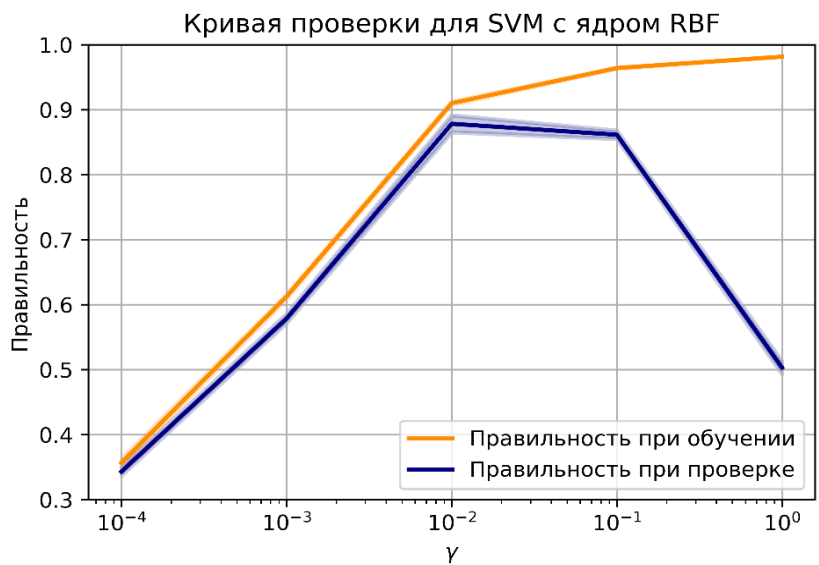

**Fig. 5.** Plot of verification curves for parameter γ at optimal

**Model training results.** The proposed algorithm, based on machine learning using the SVM method, has demonstrated its ability to correctly classify in the presence of noise typical of MEMS sensors. The best classification results were obtained using the linear and rbf kernels (verification accuracy 0.907 and 0.902, respectively). Such results are achieved in the case of the linear kernel using the optimal value of the C hyperparameter equal to 10. And in the case of using the rbf kernel, the optimal hyperparameters are: C equal to 100 and γ equal to 0.01.

**Conclusion.** SVM-based machine learning is a complex non-linear architecture that is used in many applications. In this work, we implemented an MO algorithm to improve the determination of Euler angles (roll, pitch, and yaw) from MEMS sensor data. To do this, we solved the problem of overfitting and underfitting by choosing the optimal values of hyperparameters. Further research will be directed to the application of other ML classification algorithms, as well as to increasing the types of input data.

*The work was carried out with the financial support of the Ministry of Science and Higher Education of the Russian Federation within the framework of the basic part of the state task in the field of scientific activity on the task №075-01027-21-01 from 28.09.2021.*

**REFERENCES**

1. *Shishkin Y.E. and Grekov A.N.* Koncepcija intellektual'noj sistemy avtomatizirovannogo jekologicheskogo monitoringa na baze malogabaritnyh avtonomnyh robotov (The concept of automated environmental monitoring intellectual system based on compact autonomous robots). Sistemy kontrolja okruzhajushhej sredy, 2018, No. 4(34), pp. 63–69.

2. *Shishkin Y.E., Grekov A.N., and Nikishin V.V.* Intelligent decision support system for detection of anomalies and unmanned surface vehicle inertial navigation correction. 2019 International Russian Automation Conference (RusAutoCon). IEEE, 2019, pp. 1–6.

3. *Duong D.Q., Nguyen T.P., Sun J., and Luo L.* Attitude estimation by using MEMS IMU with Fuzzy Tuned Complementary Filter. Computer Science. 2016 IEEE International Conference on Electronic Information and Communication Technology (ICEICT), 2016, DOI:10.1109/ICEICT.2016.7879720

4. *Siciliano B. and Khatib O. (eds.):* Springer Handbook of Robotics. Springer, Cham (2016). https://doi.org/10.1007/978-3-319-32552-1

5. *Malmstrom J.* Robust Navigation with GPS/INS and Adaptive Beamforming. Swedish Defence Research Agency System Technology Division SE-172 90 STOCKHOLM Sweden, 2003.

6. *Grekov A.N., Alekseev S.Y., and Bashkirov V.Y.* Rezul'taty laboratornyh ispytanij podvodnoj navigacionnoj sistemy dlja apparatov jekologicheskogo kontrolja (The results of laboratory tests underwater navigation system for environmental monitoring devices). Sistemy kontrolja okruzhajushhej sredy, 2020, No. 3 (41), pp. 65–74. DOI:10.33075/2220-5861-2020-3-65-74

7. An Introduction to Machine Learning with Python (O'Reilly) by Andreas C. Mueller and Sarah Guido. Copyright 2017 Sarah Guido and Andreas Mueller, 978-1-449-36941-5

8. *Klein I. and Asraf O.* StepNet – Deep Learning Approaches for Step Length Estimation. IEEE Access, 2020, Vol. 8, pp. 85706–85713.

9. *Jamil F., Iqbal N., Ahmad S., and Kim D.H.* Toward Accurate Position Estimation Using Learning to Prediction Algorithm in Indoor Navigation. Sensors, 2020, Vol. 20, pp. 4410.

10. *Deng J., Xu Q., Ren A., Duan Y., Zahid A., and Abbasi Q.H.* Machine Learning Driven Method for Indoor Positioning Using Inertial Measurement Unit. In Proceedings of the International Conference on UK-China Emerging Technologies (UCET), Glasgow, UK, 20-21 August 2020, pp. 1-4.

11. *Wang H.N., Yi G.X., Wang C.H., and Guan Y.* Nonlinear Initial Alignment of Strapdown Inertial Navigation System Using CSVM. In Applied Mechanics and Materials, Trans Tech Publications: StafaZurich, Switzerland, 2012, Vol. 148-149, pp. 616–620.

12. *Cortés S., Solin A., and Kannala J.* Deep learning based speed estimation for constraining strapdown inertial navigation on smartphones. In Proceedings of the IEEE 28th International Workshop on Machine Learning for Signal Processing, Aalborg, Denmark, 17–20 September 2018, pp. 1–6.

13. *Chen H., Aggarwal P., Taha T.M., and Chodavarapu V.P.* Improving Inertial Sensor by Reducing Errors using Deep Learning Methodology. In Proceedings of the NAECON 2018-IEEE National Aerospace and Electronics Conference, Dayton, OH, USA, 23–26 July 2018, pp. 197–202.

14. *Pukhov E. and Cohen H.I.* Novel Approach to Improve Performance of Inertial Navigation System Via Neural Network. In Proceedings of the 2020 IEEE/ION Position, Location and Navigation Symposium, Portland, OR, USA, 20–23 April 2020, pp. 746–754.

15. *Raschka S. and Mirjalili V.* Python Machine Learning, Third Edit. Packt, 2019, 725 p. ISBN: 978-1-78995-575-0.

16. *Wolpert D.H.* The lack of a priori distinctions between learning algorithms. Neural computation, 1996, Vol. 8, No. 7, pp. 1341–1390.

17. Scikit-learn: Machine Learning in Python, Pedregosa et al., JMLR 12, 2011, pp. 2825–2830.

18. *Witten Ian H., Frank Eibe, Hall Mark A.,and Pal Christopher J.* Data Mining: Practical Machine Learning Tools and Techniques. Amsterdam; Paris: Elsevier. 2017. https://doi.org/10.1016/C2015-0-02071-8

19. *Brunton S.L. and Kutz J.N.* Data-driven science and engineering: Machine learning, dynamical systems, and control. Cambridge University Press, 2019.

20. *Bishop C.M.* Pattern recognition. Machine learning, 2006, Vol. 128, No. 9.

21. *Smola A.J. and Schölkopf B.* A tutorial on support vector regression. Statistics and computing,

2004, Vol. 14, No. 3, pp. 199–222.

22. *Vapnik V.* The nature of statistical learning theory. Springer science & business media, 2013.

23. *Burges C.J.C.* A tutorial on support vector machines for pattern recognition. Data mining and knowledge discovery, 1998, Vol. 2, No. 2, pp. 121–167.

24. *Boser Bernhard E., Guyon Isabelle M., and Vapnik Vladimir N.* A training algorithm for optimal margin classifiers. In Proceedings of the Fifth Annual Workshop on Computational Learning Theory,

1992, ACM, pp. 144–152.

25. *Cortes C. and Vapnik V.* Support-vector networks. Machine Learning, 1995, Vol. 20 (3), pp. 273–297.

26. *Albon C.* Machine learning with python cookbook: Practical solutions from preprocessing to deep learning. O'Reilly Media, Inc., 2018.